%% file: main.tex
\documentclass[conference,10pt]{IEEEtran}

\usepackage{algpseudocode}
\algtext*{EndWhile}
\algtext*{EndIf}
\algtext*{EndFor}
\algtext*{EndProcedure}
\usepackage{amsmath}
\usepackage{amssymb}
\usepackage{array}
\usepackage{algorithm}
\usepackage{bm}
\usepackage[bookmarks=false, hidelinks]{hyperref}
\usepackage[noadjust]{cite}
\usepackage[font=small]{caption}
\usepackage{color}
\usepackage{comment}
\usepackage{dblfloatfix}
\usepackage[inline]{enumitem}
\usepackage{epsfig}
\usepackage{etoolbox}
\usepackage{fancybox}
\usepackage{float}
\usepackage{fixltx2e}
\usepackage{graphicx}
\usepackage{listings}
\usepackage{mathrsfs}
\usepackage{multirow}
\usepackage{multicol}
\usepackage{pifont}
\usepackage{placeins}
\usepackage{rotating}
\usepackage{setspace}
\usepackage{soul}
\usepackage[hang, tight]{subfigure}
\usepackage{threeparttable}
\usepackage{times}
\usepackage{url}
\usepackage{upgreek}
\usepackage{verbatim}
\usepackage{wrapfig}
\usepackage[usenames,dvipsnames]{xcolor}
\usepackage{authblk}
\usepackage[]{siunitx}
\usepackage[super]{nth}
\usepackage{tabularx}
\usepackage{supertabular}
\usepackage{tikz}
\usepackage{bbm}
\usepackage{blindtext}
\usepackage{booktabs}
\usepackage{diagbox}

\usepackage{tablefootnote}

\usepackage{tikz}
\usepackage{xcolor}
\newcommand*\circled[1]{\tikz[baseline=(char.base)]{
            \node[shape=circle,fill,inner sep=1pt,scale=0.8] (char) {\textcolor{white}{#1}};}}

\usepackage{listings}
\usepackage{xcolor}

\definecolor{codegreen}{rgb}{0,0.6,0}
\definecolor{codegray}{rgb}{0.5,0.5,0.5}
\definecolor{codepurple}{rgb}{0.58,0,0.82}
\definecolor{backcolour}{rgb}{0.95,0.95,0.92}

\lstdefinestyle{mystyle}{
  backgroundcolor=\color{backcolour},   commentstyle=\color{codegreen},
  keywordstyle=\color{magenta},
  numberstyle=\tiny\color{codegray},
  stringstyle=\color{codepurple},
  basicstyle=\ttfamily\footnotesize,
  breakatwhitespace=false,         
  breaklines=true,                 
  captionpos=b,                    
  keepspaces=true,                 
  numbers=left,                    
  numbersep=5pt,                  
  showspaces=false,                
  showstringspaces=false,
  showtabs=false,                  
  tabsize=2
}
\usepackage{colortbl}
\usepackage{dcolumn}

\definecolor{greenDeep}{RGB}{0,170,0}
\definecolor{greenSlightDeep}{RGB}{0,205,0}
\definecolor{greenShallow}{RGB}{0,255,0}
\definecolor{greenShallower}{RGB}{160,255,0}
\definecolor{orangeShallow}{RGB}{255,190,0}
\definecolor{orangeDeep}{RGB}{255,80,0}
\definecolor{orangeDeeper}{RGB}{255,40,0}
\definecolor{redDeep}{RGB}{255,0,0}

\definecolor{redLight}{RGB}{255,128,114}

\def\zz#1{%
\ifdim#1pt>4.9pt\cellcolor{greenDeep}\else
\ifdim#1pt>3.9pt\cellcolor{greenSlightDeep}\else
\ifdim#1pt>2.9pt\cellcolor{greenShallower}\else
\ifdim#1pt>2.9pt\cellcolor{yellow}\else
\ifdim#1pt>1.9pt\cellcolor{orangeShallow}\else
\ifdim#1pt>1.9pt\cellcolor{orange}\else
\ifdim#1pt>0.9pt\cellcolor{orange}\else
\ifdim#1pt>0.9pt\cellcolor{orangeDeep}\else
\cellcolor{orangeDeep}\fi\fi\fi\fi\fi\fi\fi\fi
#1}

\def\correct#1{%
\cellcolor{greenSlightDeep}
#1}

\def\wrong#1{%
\cellcolor{redLight}
#1}

\graphicspath{{./_fig/}}

\makeatletter
\renewcommand\footnoterule{%
  \kern-3\p@
  \hrule\@width0.4\columnwidth
  \kern2.6\p@}
  \makeatother

\usepackage[hang,flushmargin]{footmisc}

\begin{document}






\title{\vspace{-.1in}RTLLM: An Open-Source Benchmark for Design \underline{RTL} Generation with \underline{L}arge \underline{L}anguage \underline{M}odel\vspace{-.1in}}

\author[]{ \fontsize{12}{12}\selectfont Yao Lu$^*$, Shang Liu$^*$, Qijun Zhang, Zhiyao Xie\textsuperscript{$\dagger$}\vspace{-6pt}}

\affil[]{\fontsize{10}{10}\selectfont Hong Kong University of Science and Technology\vspace{-7pt}}


\affil[]{\{yludf, sliudx, qzhangcs\}@connect.ust.hk, eezhiyao@ust.hk\vspace{-12pt}}


\maketitle

\begingroup\renewcommand\thefootnote{$*$}
\footnotetext{Equal Contribution}
\endgroup

\begingroup\renewcommand\thefootnote{$\dagger$}
\footnotetext{Corresponding Author}
\endgroup

\input{_txt/abstract}

\input{_txt/1_introduction}

\input{_txt/2_preliminaries}

\input{_txt/4_Methodology}

\input{_txt/5_experiment}

\input{_txt/6_discussion}
\input{_txt/6_conclusion}

\bibliographystyle{IEEEtran}
\bibliography{references_1, references_2}
\end{document}

%% file: _txt/abstract.tex
\begin{abstract}



Inspired by the recent success of large language models (LLMs) like ChatGPT, researchers start to explore the adoption of LLMs for agile hardware design, such as generating design RTL based on natural-language instructions. However, in existing works, their target designs are all relatively simple and in a small scale, and proposed by the authors themselves, making a fair comparison among different LLM solutions challenging. In addition, many prior works only focus on the design correctness, without evaluating the design qualities of generated design RTL. In this work, we propose an open-source benchmark named RTLLM, for generating design RTL with natural language instructions. To systematically evaluate the auto-generated design RTL, we summarized three progressive goals, named syntax goal, functionality goal, and design quality goal. This benchmark can automatically provide a quantitative evaluation of any given LLM-based solution. Furthermore, we propose an easy-to-use yet surprisingly effective prompt engineering technique named self-planning, which proves to significantly boost the performance of GPT-3.5 in our proposed benchmark.

    
\end{abstract}

%% file: _txt/1_introduction.tex
\section{Introduction}

In recent years, machine learning (ML) for EDA, or named ML for circuit/hardware design, has become a trending topic~\cite{huang2021machine, rapp2021mlcad}. 
By learning from prior design solutions, ML models can perform fast circuit quality evaluations or even optimizations. Most existing ML for EDA solutions can be categorized into two main types, \emph{predictive} models and \emph{generative} models. 
\emph{Predictive} ML models are trained to provide early predictions on circuit qualities. 
In contrast, \emph{generative} models are supposed to generate design solutions directly, which is more useful while challenging.


Recently, natural language processing (NLP) researchers realize that when the scale of model parameters exceeds a certain level, these enlarged language models can achieve a significant performance improvement over small-scale language models like BERT~\cite{devlin2018bert}. The most remarkable progress of large language models (LLMs) is reflected by the popularity of commercial products GPT-3.5 and GPT-4~\cite{gpt4}.


Inspired by this recent success of LLMs, researchers start to explore the adoption of LLMs for agile hardware design. One intuitive and promising direction is to generate the target design RTL directly with natural language instructions. This new paradigm is expected to significantly reduce the barrier of hardware design and improve the design productivity. 
Such natural-language-based design method may revolutionize existing design methods based on hardware description language (HDL), including Verilog, VHDL, Chisel, C++/SystemC with high-level synthesis (HLS), etc. 


There have been some most recent explorations~\cite{thakur2023benchmarking, blocklove2023chip, chang2023chipgpt} on this topic. Thakur et al. proposes to fine-tune open-source LLMs like CodeGen~\cite{nijkamp2022conversational} to generate Verilog code for target designs~\cite{thakur2023benchmarking}. Then Chip-Chat~\cite{blocklove2023chip} further discusses the challenges and opportunities in hardware design based on LLMs. It indicates an obviously superior performance of ChatGPT over open-sourced LLMs. Another work Chip-GPT~\cite{chang2023chipgpt} studies a similar task, proposing to perform RTL design based on ChatGPT. We expect more explorations in natural-language based hardware design based on LLMs in the future.

\begin{table}[!t]
      \centering
      \vspace{.05in}
     \hspace{-.15in}
      \renewcommand{\arraystretch}{1.1}
      \resizebox{0.49\textwidth}{!}{
        \begin{tabular}{ |c|c||c|c| } 
        \hline
        \multirow{2}{*}{Works}  &  Num of  &  Num of HDL Lines  & Num of Cells in Netlist\tablefootnote{Excluding the pseudo RAM design implemented with a large matrix of D flip-flop, because its complexity completely depends on the number of wordlines and bitlines as parameters. Also, realistic SRAMs should consist of SRAM cells instead of flip-flops and be generated by memory compilers.}   \\
        \cline{3-4}
                     &   Designs     &  \multicolumn{2} {c|} {\{Medium, Mean, Max, Total\}} \\ 
        \hline
         \hline
         Thakur et al.~\cite{thakur2023benchmarking} &  17  & \{16, 19, 48, 0.3K\} & \{9.5, 45, 335, 0.7K\}  \\
         \hline
         Chip-Chat~\cite{blocklove2023chip} &  8   &   \{42, 42, 72, 0.3K\}  &  \{37, 44, 110, 0.4K\} \\
         \hline
         Chip-GPT~\cite{chang2023chipgpt}  &  8   &  \multicolumn{2} {c|} {Not released to public}  \\
         \hline
         \hline
         RTLLM   &  30   &  \{52, 86, 518, 2.5K\} &   \{121, 408, 2435, 11.8K\}   \\
         \hline
        \end{tabular}
       }
        \caption{The statistics of designs evaluated in prior works~\cite{thakur2023benchmarking, blocklove2023chip, chang2023chipgpt} and in RTLLM. We quantify the design complexity with the number of HDL lines in each design RTL and the design scale with the number of cells in the post-synthesis netlist. RTLLM is an obviously more comprehensive benchmark compared with other datasets.}
        \label{resource_table}
        \vspace{-.2in}
\end{table}

However, in these existing works~\cite{thakur2023benchmarking,blocklove2023chip,chang2023chipgpt}, their target designs are all relatively simple and in a small circuit scale, as shown in Table~\ref{resource_table}. As a result, the performance and scalability of LLM solutions are not thoroughly evaluated. In addition, these small designs are proposed by the authors themselves, making a fair comparison among different LLM solutions challenging. More importantly, even for the same design, the natural language description from different human designers can be largely different. Using a unified natural language design description is necessary for fair LLM evaluations.

In this work, we propose a comprehensive open-source benchmark for design RTL generation with natural language. It is named RTLLM. It supports the evaluation of any generated HDL format, including Verilog, VHDL, and Chisel, as long as it supports logic synthesis and RTL simulation. RTLLM consists of 30 designs with a wide coverage of design complexities and scales. To systematically and quantitatively evaluate the quality of each auto-generated design RTL, we summarize three progressive goals, named syntax goal, functionality goal, and design quality goal. Based on our provided design automation scripts, the benchmark can automatically evaluate any given LLM solution with respect to all three goals. More importantly, RTLLM provides ground-truth design RTLs crafted by human designers, providing a standard baseline to evaluate the design quality goal. 

Our contributions in this work are summarized below.

\begin{itemize}
\item We propose a comprehensive open-source benchmark\footnote{It will be open-sourced in https://github.com/hkust-zhiyao/RTLLM} dedicated to the automatic design RTL generation with natural languages. Compared with the released dataset in recent works, our benchmark includes many more designs, also with higher design scale and complexity. 
\item We systematically evaluated state-of-the-art commercial and academic solutions with our benchmark. In addition to assessing syntax and functionality, we also evaluate the design PPAs of the generated RTL by comparing it with our human-crafted designs provided in RTLLM.
\item Besides providing the benchmark, we also propose an innovative new prompt engineering technique named self-planning, without requiring any human interference. Combining self-planning and GPT-3.5 can well outperform the performance of GPT-3.5 and get close to GPT-4's state-of-the-art performance.  
\end{itemize}

%% file: _txt/2_preliminaries.tex
\section{Problem Formulation}

In this section, we provide a general formulation of the RTL generation task based on natural language instructions. Given a natural language description of desired design functionality named $\mathcal{L}$, the target is to develop an ML model $F$ to generate the RTL of this design $\mathcal{V}$, with $\mathcal{V} = F(\mathcal{L})$. To achieve this goal, currently the model $F$ is based on LLMs. 


However, the generation directly based on the LLM $F$  may not be successful. Therefore, prompt engineering techniques $P$ can be applied to revise the design functionality description in natural language $\mathcal{L}$, generating $\mathcal{L}_P = P(\mathcal{L})$, which is feed into LLMs $F$ as input. In addition, this LLM output may be further manually revised by human engineers $H$, making the ultimate output $\mathcal{V} = H(F(\mathcal{L}_P))$.   


%% file: _txt/4_Methodology.tex
\linespread{1.2}
\begin{table*}[!th]
\caption{Benchmark Descriptions and Scales}
\vspace{-.1in}
\begin{center}
\label{benchmark_detail}
      \resizebox{0.97\textwidth}{!}{
\begin{tabular}{cllcc}
\toprule[1pt]
  & \textbf{Design} & 
  \textbf{Description} &
  \textbf{\begin{tabular}[c]{@{}c@{}}Lines of \\ Code\end{tabular}} &
  \textbf{\begin{tabular}[c]{@{}c@{}}Circuit Scale\\ (Cells)\end{tabular}} \\
\midrule[0.6pt]
\multirow{7}{*}{Arithmetic} & accu              & Accumulates 8-bit data and output after 4 inputs                                & 64  & 195  \\
                            & adder\_8bit       & An 8-bit adder                                                     & 26  & 58   \\
                            & adder\_16bit      & A 16-bit adder implemented with full adders                                                            & 137 & 130  \\
                            & adder\_32bit      & A 32-bit carry-lookahead adder                                                  & 181 & 312  \\
                            & adder\_64bit      & A 64-bit ripple carry adder based on 4-stage pipeline                                                 & 197 & 1340  \\
                            & multi\_8bit       & An 8-bit booth-4 multiplier                                                     & 84  & 34   \\
                            & multi\_16bit      & An 16-bit multiplier based on shifting and adding operation                     & 65  & 817   \\
                            & multi\_pipe\_4bit & A 4-bit unsigned number pipeline multiplier                                     & 43  & 120  \\
                            & multi\_pipe\_8bit & An 8-bit unsigned number pipeline multiplier                                    & 92  & 578  \\
                            & div\_8bit         & An 8-bit radix-2 divider                                                        & 72  & 94   \\
                            & div\_16bit        & A 16-bit divider based on subtraction operation                                & 45  & 1855   \\
\midrule[0.6pt]
\multirow{17}{*}{Logic}     & JC\_counter       & 4-bit Johnson counter with specific cyclic state sequence                       & 22  & 134  \\
                            & right\_shifter    & Right shifter with 8-bit delay                                                  & 17  & 466  \\
                            & mux               & Multi-bit mux synchronizer                                                      & 46  & 19   \\
                            & counter\_12       & Counter module counts from 0 to 12                                              & 37  & 38   \\
                            & freq\_div         & Frequency divider for 100M input clock, outputs 50MHz, 10MHz, 1MHz              & 51  & 64   \\
                            & signal\_generator & Signal generator produces square, sawtooth, and triangular waveforms            & 52  & 135  \\
                            & serial2parallel   & 1-bit serial input and output data after receiving 6 inputs                     & 62  & 66   \\
                            & parallel2serial   & Convert 4 input bits to 1 output bit                                            & 41  & 24   \\
                            & pulse\_detect     & Extract pulse signal from the fast clock and create a new one in the slow clock & 38  & 6    \\
                            & edge\_detect      & Detect rising and falling edges of changing 1-bit signal                        & 39  & 7    \\
                            & FSM               & FSM detection circuit for specific input                                        & 77  & 24   \\
                            & width\_8to16      & First 8-bit data placed in higher 8-bits of the 16-bit output                   & 50  & 117  \\
                            & traffic\_light    & Traffic light system with three colors and pedestrian button                    & 106 & 117  \\
                            & calendar          & Perpetual calendar with seconds, minutes, and hours                             & 37  & 121  \\
                            & RAM               & 8x4 bits true dual-port RAM                                                     & 50  & 1834 \\
                            & asyn\_fifo        & An asynchronous FIFO 16×8 bits                                                  & 149 & 686  \\
                            & ALU               & An ALU for 32bit MIPS-ISA CPU                                                   & 111 & 2435 \\
                            & PE                & A Multiplying Accumulator for 32bit integer                                     & 27  & 1439 \\
                            & risc\_cpu         &\begin{tabular}[c]{@{}l@{}}Simplified RISC\_CPU with clock generator, instruction register, accumulator, \\ arithmetic logic unit, data controller, state controller, etc.\end{tabular}                           & 518 & 407 \\ 
\bottomrule[1pt] 
\end{tabular}
}
\end{center}
\vspace{-.2in}
\end{table*}

\begin{figure}[!t]
\includegraphics[width=0.48\textwidth]{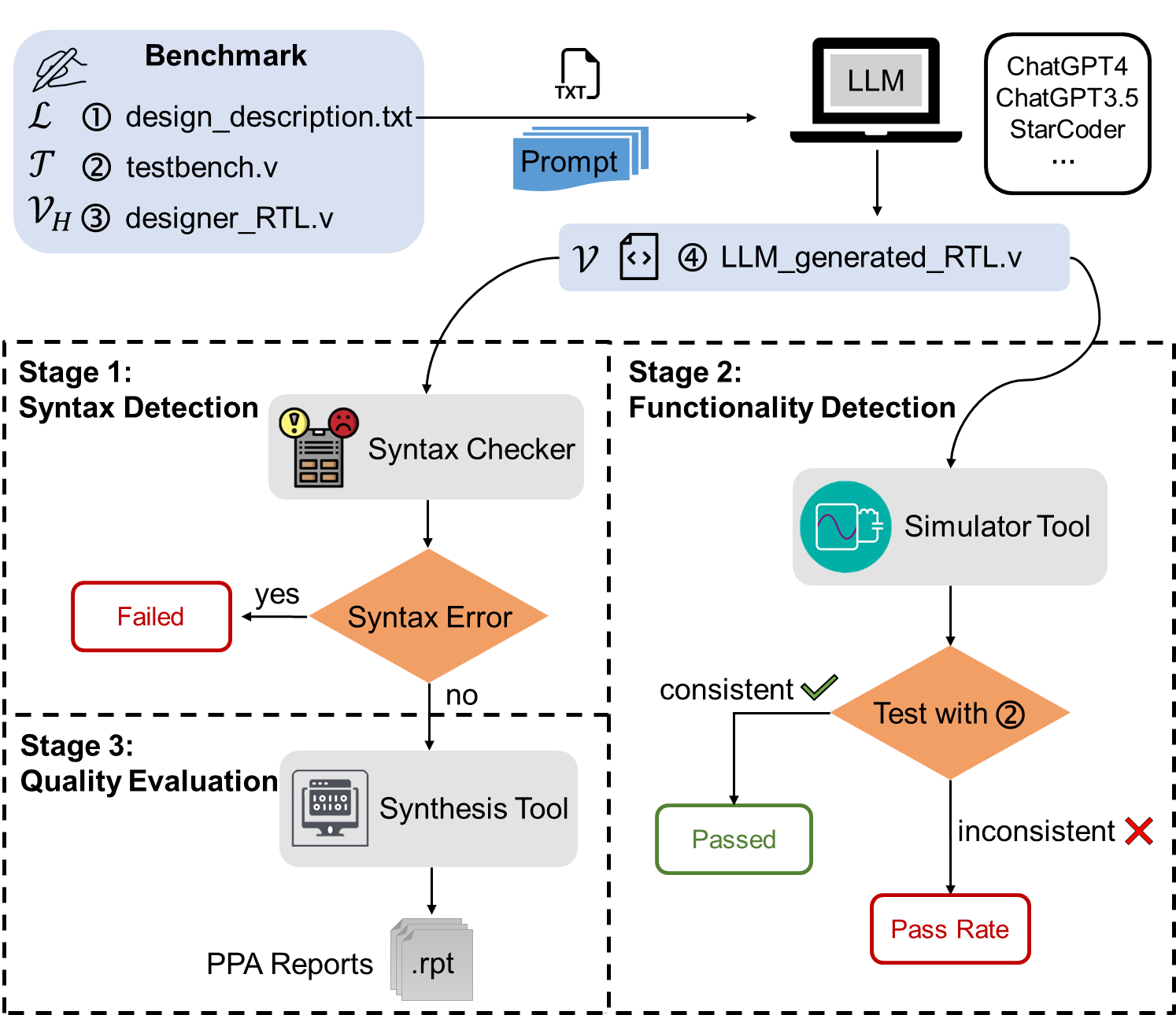}
\caption{The workflow of adopting RTLLM for completely automated design RTL generation and evaluation. The user only needs to provide their LLM as input. It evaluates whether each generated design satisfies the syntax goal, functionality goal, and quality goal.}
\label{benchmark}
\vspace{-.2in}
\end{figure}

\section{RTLLM: An RTL Generation Benchmark} \label{sec:algorithm}

\subsection{Evaluation Metrics for RTL Generation Task}

To systematically evaluate the generated design RTL $\mathcal{V}$, we summarize three progressive goals of the $\mathcal{V}$. Our benchmark enables automatic evaluation of these three goals as three metrics. These goals are summarized as below.  

We name the first and the most fundamental goal as the \textbf{syntax goal}. It means the syntax of generated RTL design $\mathcal{V}$ should at least be correct. It can be verified by checking whether the design can be correctly synthesized into netlist by synthesis tools~\cite{design-compilier} without syntax errors. 

After ensuring syntax correctness, we name the second goal as the \textbf{functionality goal}. It means the functionality of generated RTL design $\mathcal{V}$ should be exactly the same as the designers' expectation. It can be verified by checking whether the generated design passes all test cases in a comprehensive testbench. Of course, exhausting all possible test cases will make the testbench file extremely cumbersome. Our benchmark only samples a reasonable number of test cases. Passing all test cases does not necessarily mean the functionality is 100\% correct.

If the generated design RTL $\mathcal{V}$ proves correct in both syntax and functionality, the design can be viewed as successful. But in order to make $\mathcal{V}$ practically useful, its design qualities including performance, power, area (PPA) should also be desirable. We name this goal as \textbf{quality goal}. It can be verified by measuring the PPA values after the synthesis and layout of generated $\mathcal{V}$. This quality goal is not explicitly evaluated in prior works~\cite{thakur2023benchmarking,blocklove2023chip,chang2023chipgpt}.

\subsection{An Overview of the Design Generation Benchmark}

RTLLM collects 30 common designs with various design scales and complexities. For each design, the benchmark provides the following information in three separate files. 
\begin{itemize}
    \item \textbf{Description} (\emph{design\_description.txt}) denoted as $\mathcal{L}$: A natural language description of the target design's functionality. The criterion is that a human designer can accurately write a correct design RTL $\mathcal{V}$ after reading the description $\mathcal{L}$. This description $\mathcal{L}$ also includes an explicit indication of the module name, all input and output (I/O) signals with signal name and width. These pre-defined modules and I/O signal information enable automatic functionality verification with our provided testbench.  
    \item \textbf{Testbench} (\emph{testbench.v}) denoted as $\mathcal{T}$: A testbench with multiple test cases, each with input values and correct output values. The testbench corresponds to the pre-defined module name and I/O signals in $\mathcal{L}$. It can be applied to verify the correctness of design functionality. 
    \item \textbf{Correct Design} (\emph{designer\_RTL.v}) denoted as $\mathcal{V}_H$: A reference design Verilog hand-crafted by human designers. By comparing with this reference design $\mathcal{V}_H$, we can quantitatively evaluate the design qualities of the automatically generated design $\mathcal{V}$. Also, these correct designs have all passed our proposed testbenches. 
\end{itemize}

Fig.~\ref{benchmark} shows a complete workflow of RTL generation and evaluation using this benchmark, including three straightforward stages. In stage \circled{1}, users feed each natural language description $\mathcal{L}$ into their target LLM $F$, generating the design RTL $\mathcal{V} = F(\mathcal{L})$. 
If an LLM solution requires additional prompt techniques $P$, it will switch the natural language description $\mathcal{L}$ to actual input prompts $\mathcal{L}_P$, with the output design RTL being $\mathcal{V} = F(\mathcal{L}_P)$. If necessary, additional human engineers' efforts can also be introduced, generating $\mathcal{V} = H(F(\mathcal{L}_P))$.

In stage \circled{2}, the framework tests the functionality of generated design RTL $\mathcal{V}$ using our provided testbench $\mathcal{T}$. In stage \circled{3}, the generated design RTL $\mathcal{V}$ is synthesized into a netlist to analyze the design qualities in terms of PPA values, which are compared with the design qualities of the provided reference designs $\mathcal{V}_H$. This whole process is automated.



\subsection{Detailed Inspection of the Benchmark}

Table~\ref{benchmark_detail} shows the detailed description of all 30 designs in our provided benchmark. It provides 30 common digital designs, with 11 arithmetic designs and the other 19 logic designs implementing various functionalities. All reference designs from human engineers $\mathcal{V}_H$ are coded in Verilog. To give more information about the design complexity and design scale, Table~\ref{benchmark_detail} also provides the number of lines in the HDL code and the number of cells in the synthesized gate-level netlist. Intuitively, a design with more HDL code tends to be more complex to implement by human designers, and a design with more cells in netlist is naturally larger in the design area and power consumption. These statistics are collected based on the correct reference designs $\mathcal{V}_H$ from designers.


For the 11 arithmetic designs in RTLLM, they cover the most common design types including accumulators, adders, multipliers, and dividers, with all common bit widths from 4 bits to 64 bits. For each design type, we cover different implementation requirements. Take the adder as an example, the benchmark includes the basic version without any requirement (i.e., adder\_8bit), the adder implemented with 1-bit full adders (i.e., add\_16bit), the lookahead adder (i.e., adder\_32bit), and the ripple adder with pipelines (i.e., adder\_64bit). Both design complexity and scale increase progressively in these adders.

The 19 logic designs in RTLLM include designs with more variations in their target functionalities. It includes simpler designs like counters (i.e., counter\_12) and finite state machines (i.e., FSM), and more complex designs like the simplified RISC CPU design (i.e., risc\_cpu) and a processing element (i.e., PE) performing multiply–accumulate operations.

In summary, RTLLM proposes 30 common designs with rich diversities in their functionalities, implementation requirements, design complexities, and design scales. The overall scale of RTLLM is significantly larger than the data released in prior works~\cite{thakur2023benchmarking, blocklove2023chip}, as already summarized in Table~\ref{resource_table}.

\section{Self-Planning Technique}


To further explore the capabilities of LLMs, in this work, we also propose a highly effective prompt engineering technique named self-planning. It is extremely easy to use and works surprisingly well. Instead of directly generating design RTL in one query, self-planning decomposes this enquiry into a two-step process, without requiring any extra efforts from human users or any existing design data.

\begin{lstlisting}[language=Verilog, caption={Part of the input of the first step in self-planning. Besides the design description with the definition of module I/O, it requests the LLM to first generate the reasoning steps (line 5) and advices to avoid errors (line 6).}, linewidth=.48\textwidth, float=tp, belowskip=-2\baselineskip, label={lst:planning}]
#Implement the design of unsigned 16bit multiplier based on shifting and adding opration.  
module multi_16bit(
// ...I/O details omitted...
);
#Please act as a professional verilog designer, try to understand the requirements above and give reasoning steps in natural language to achieve it. 
#In addition, try to give advice to avoid syntax error.
\end{lstlisting}

\begin{lstlisting}[language=Verilog, caption={Part of the output of the first step from self-planning.}, linewidth=.48\textwidth, float=b, aboveskip=-1\baselineskip, belowskip=-1\baselineskip, label={lst:planning_out}]
The task is to implement ... The modules has an ...
To achieve this, we will use a state machine to control the multiplication process. The steps involved are as follows:
IDLE: The initial state where the module waits for the start signal to be asserted.
MULTIPLY: The state where the multiplication operation is performed.
FINISH: The final state indicating the multiplication operation is complete.
Create the necessary registers to hold the state, the inputs (ain and bin), and the intermediate results during the multiplication process.
// ...details omitted...
Potential Syntax Error Advice:
Ensure that all the signals ... are declared ...
Make sure to use blocking ... non-blocking ... 
Ensure that you use the correct parameter size for states. Since we have three states, we need at least two bits to represent them ...
Check for any typos or misspellings ...
\end{lstlisting}

 
The first step requests the LLM to \emph{plan} how to write the target design $\mathcal{V}$. Specifically, the model is required to output natural language-level analysis and reasoning steps of the target task. Additionally, considering that language models sometimes overlook the syntax requirements in Verilog generation, we also require the model to provide the syntax errors it needs to avoid during the code generation process. The LLM outputs are collected for the second step. 


Code.~\ref{lst:planning} shows a partial input example in the first step of self-planning when generating the multi\_16bit design. Lines 5 and 6 at the end give the planning instructions. Part of the LLM output is shown in Code.~\ref{lst:planning_out}. As requested, the output in Code.~\ref{lst:planning_out} includes both a detailed plan with reasoning steps, and advice to avoid potential syntax errors.

In the second step, both the original design description $\mathcal{L}$ and the \emph{plan} and \emph{advice} collected in the first step will be fed into the LLM to generate the final design RTL $\mathcal{V}$. Compared with directly asking for the result, the design RTL generated by this two-step self-planning is obviously less error-prone.


Code~\ref{lst:result_multi} and~\ref{lst:result_add} present the wrong RTL designs generated by GPT-3.5 without using self-planning, targeting generating multi\_16bit and adder\_32bit respectively. After using self-planning with GPT-3.5, both the syntax error in Code~\ref{lst:result_multi} and the functionality error in Code~\ref{lst:result_add} are solved, generating correct design RTL for both designs. 

\begin{lstlisting}[language=Verilog, caption={A wrong design RTL of multi\_16bit, generated by GPT-3.5 without self-planning. There are syntax errors in this design. It wrongly defines new variables (i.e., \textbf{i}, \textbf{a\_shifted}) in the always block. After adopting self-planning, GPT-3.5 can generate correct design, likely benefiting from the syntax error advice  from the first step.}, linewidth=.48\textwidth, float=tp, belowskip=-2\baselineskip, label={lst:result_multi}]
module multi_16bit(
// ...I/O details omitted...
);
// ...details omitted...
always @(posedge clk or negedge rst_n) begin
    integer i;
    reg [15:0] a_shifted; 
    for (i = 0; i < 16; i = i + 1) begin
        // ......
    end
end
// ......
\end{lstlisting}

\begin{lstlisting}[language=Verilog, caption={A wrong design RTL of adder\_32bit, generated by GPT-3.5 without self-planning. Despite correct syntax, the functionality of this design is wrong, epecially reflected in its usage of the \textbf{carry} variable. After adopting self-planning, GPT-3.5 can generate correct design, likely benefiting from the reasoning steps from the first step.}, linewidth=.48\textwidth, float=b, belowskip=-1\baselineskip, aboveskip=-1\baselineskip, label={lst:result_add}]
module adder_32bit(
// ...I/O details omitted...
);
// ...details omitted...
    for (i = 1; i <= 32; i = i + 1) begin : cla_block
        wire P, G, carry;
        //......
        assign carry = (i == 1) ? 1'b0 : (P & {carry, G});
        assign S[i] = A[i] ^ B[i] ^ {carry, G};
        //......
    end
end
\end{lstlisting}

\linespread{1.2}
\begin{table*}[htbp]
\vspace{-.2in}
\caption{The Syntax and Functionality Correctness Verification for Different LLMs}
\label{syntax_function}
\vspace{-.1in}
\begin{center}
\resizebox{0.83\textwidth}{!}{
\begin{tabular}{l|cc|cc|cc|cc|cc|cc}
\toprule[1pt]
\multirow{2}{*}{\textbf{Design}} &
  \multicolumn{2}{c|}{\textbf{GPT-3.5}} &
  \multicolumn{2}{c|}{\textbf{GPT-4}} &
  \multicolumn{2}{c|}{\textbf{Thakur et al.~\cite{thakur2023benchmarking}}} &
  \multicolumn{2}{c|}{\textbf{StarCoder~\cite{li2023starcoder}}} &
  \multicolumn{2}{c|}{\textbf{GPT-3.5 + SP}}&
  \multicolumn{2}{c}{\textbf{GPT-4.0 + SP}}\\ 
  &
  Syntax & Func. & Syntax & Func. & Syntax & Func. & Syntax & Func. & Syntax &
  Func. & Syntax & Func.\\\midrule[0.6pt]
accu              & \textbf{\zz{4}} & \ding{52}  & \textbf{\zz{5}} & \ding{52}  & \textbf{\zz{0}} & - & \textbf{\zz{0}} & - & \textbf{\zz{4}} & \ding{52}& \textbf{\zz{5}} & \ding{52} \\
adder\_8bit       & \textbf{\zz{4}} & \ding{52} & \textbf{\zz{5}} & \ding{52} & \textbf{\zz{0}} & -  & \textbf{\zz{0}} & - & \textbf{\zz{4}} & \ding{52}& \textbf{\zz{5}} & \ding{52}\\
adder\_16bit      & \textbf{\zz{5}} & \ding{56}  & \textbf{\zz{5}} & \ding{52} & \textbf{\zz{5}} &  \ding{52}   & \textbf{\zz{0}} & - & \textbf{\zz{5}} & \ding{52}& \textbf{\zz{5}} & \ding{52}\\
adder\_32bit      & \textbf{\zz{5}} & \ding{56} & \textbf{\zz{5}} & \ding{56} & \textbf{\zz{0}} & - & \textbf{\zz{0}} & - & \textbf{\zz{5}} & \ding{52}& \textbf{\zz{5}} & \ding{56} \\
adder\_64bit      & \textbf{\zz{2}} & \ding{56} & \textbf{\zz{3}} & \ding{56} & \textbf{\zz{0}}  & -  &  \textbf{\zz{0}} & - & \textbf{\zz{4}} & \ding{56}& \textbf{\zz{5}} & \ding{56}\\
multi\_8bit       & \textbf{\zz{3}} & \ding{56}  & \textbf{\zz{4}} & \ding{56} & \textbf{\zz{0}}  & -  &  \textbf{\zz{0}} &  - & \textbf{\zz{5}} & \ding{56}& \textbf{\zz{5}} & \ding{56} \\
multi\_16bit      & \textbf{\zz{0}} & - & \textbf{\zz{5}} & \ding{52} & \textbf{\zz{5}}  & \ding{52}  & \textbf{\zz{0}}  & -  & \textbf{\zz{2}} & \ding{52}& \textbf{\zz{2}} & \ding{52}\\
multi\_pipe\_4bit & \textbf{\zz{0}} & - & \textbf{\zz{2}} &  \ding{52} & \textbf{\zz{0}}  & -  &  \textbf{\zz{5}}  &\ding{52} & \textbf{\zz{1}} & \ding{56}& \textbf{\zz{5}} & \ding{56}\\

multi\_pipe\_8bit & \textbf{\zz{0}} & -           & \textbf{\zz{4}} & \ding{56} & \textbf{\zz{0}} & -         &  \textbf{\zz{5}} & \ding{52} &\textbf{\zz{3}} & \ding{56}& \textbf{\zz{4}} & \ding{52}\\
div\_8bit         & \textbf{\zz{0}} & -           & \textbf{\zz{0}} & -         & \textbf{\zz{0}} & -         & \textbf{\zz{0}} & - &\textbf{\zz{4}} & \ding{56}& \textbf{\zz{0}} & -\\
div\_16bit        & \textbf{\zz{0}} & -           & \textbf{\zz{5}} & \ding{56} & \textbf{\zz{0}} & -         & \textbf{\zz{0}}  & -  &\textbf{\zz{0}} & -& \textbf{\zz{5}} & \ding{56}\\\midrule[0.6pt]

JC\_counter       & \textbf{\zz{5}} & \ding{52} & \textbf{\zz{5}} & \ding{52} & \textbf{\zz{5}} & \ding{56}  & \textbf{\zz{5}} & \ding{56}  & \textbf{\zz{4}} & \ding{52}& \textbf{\zz{5}} & \ding{52}\\
right\_shifter    & \textbf{\zz{5}} & \ding{52} & \textbf{\zz{5}} & \ding{52} & \textbf{\zz{5}} & \ding{52}  & \textbf{\zz{0}} & -&\textbf{\zz{5}} & \ding{52}& \textbf{\zz{5}} & \ding{52}\\
mux               & \textbf{\zz{0}} & - & \textbf{\zz{4}} & \ding{52}  & \textbf{\zz{5}} & \ding{52}  & \textbf{\zz{0}} & - & \textbf{\zz{4}} & \ding{56}& \textbf{\zz{5}} & \ding{52}\\
counter\_12       & \textbf{\zz{5}} & \ding{52} & \textbf{\zz{5}} & \ding{52}  & \textbf{\zz{5}} & \ding{56}  & \textbf{\zz{5}} & \ding{56} & \textbf{\zz{4}} & \ding{52}& \textbf{\zz{5}} & \ding{52}\\
freq\_div         & \textbf{\zz{5}} & \ding{56} & \textbf{\zz{5}} & \ding{56} & \textbf{\zz{5}} & \ding{56}  & \textbf{\zz{0}} & -  & \textbf{\zz{5}} & \ding{56}& \textbf{\zz{5}} & \ding{52}\\
signal\_generator & \textbf{\zz{5}} &\ding{52}& \textbf{\zz{5}} & \ding{52}  & \textbf{\zz{0}} & - & \textbf{\zz{5}} & \ding{56} & \textbf{\zz{5}} & \ding{52}& \textbf{\zz{5}} & \ding{52}\\
serial2parallel   & \textbf{\zz{5}} & \ding{52}  & \textbf{\zz{5}} & \ding{52}  & \textbf{\zz{0}} & - & \textbf{\zz{5}} & \ding{56} & \textbf{\zz{5}} & \ding{52}& \textbf{\zz{4}} & \ding{52}\\
parallel2serial   & \textbf{\zz{5}} & \ding{56}  & \textbf{\zz{4}} &  \ding{56} & \textbf{\zz{0}} & - & \textbf{\zz{0}} & - & \textbf{\zz{3}} & \ding{56}& \textbf{\zz{5}} & \ding{56}\\
pulse\_detect     & \textbf{\zz{1}} & \ding{56} & \textbf{\zz{5}} & \ding{56}  & \textbf{\zz{5}} & \ding{56}  & \textbf{\zz{0}} & -  & \textbf{\zz{5}} & \ding{56} & \textbf{\zz{5}} & \ding{56}\\
edge\_detect      & \textbf{\zz{5}} & \ding{52} & \textbf{\zz{5}} & \ding{52}  & \textbf{\zz{5}} & \ding{56}  & \textbf{\zz{0}} &  - & \textbf{\zz{5}} & \ding{52}& \textbf{\zz{5}} & \ding{52}\\
FSM               & \textbf{\zz{5}} & \ding{56} & \textbf{\zz{5}} & \ding{56} & \textbf{\zz{5}} & \ding{56}  & \textbf{\zz{0}} & -  & \textbf{\zz{5}} & \ding{56}& \textbf{\zz{5}} & \ding{56}\\
width\_8to16      & \textbf{\zz{5}} & \ding{52}  & \textbf{\zz{5}} & \ding{52}  & \textbf{\zz{0}} & -  & \textbf{\zz{5}} & \ding{52} & \textbf{\zz{5}} & \ding{52}& \textbf{\zz{5}} & \ding{52}\\
traffic\_light    & \textbf{\zz{5}} & \ding{56} & \textbf{\zz{5}} & \ding{52} & \textbf{\zz{0}} & - & \textbf{\zz{0}} & - & \textbf{\zz{5}} & \ding{56} & \textbf{\zz{5}} & \ding{52}\\
calendar          & \textbf{\zz{0}} & - & \textbf{\zz{5}} & \ding{56}  & \textbf{\zz{0}} & - & \textbf{\zz{0}} & -  & \textbf{\zz{5}} & \ding{52}& \textbf{\zz{5}} & \ding{52}\\
RAM               & \textbf{\zz{0}} & - & \textbf{\zz{0}} & - & \textbf{\zz{5}} &  \ding{52}  & \textbf{\zz{5}} & \ding{52} & \textbf{\zz{0}} & -& \textbf{\zz{3}} & \ding{52}\\
asyn\_fifo        & \textbf{\zz{0}} & - & \textbf{\zz{0}} & - & \textbf{\zz{0}} & - & \textbf{\zz{0}} & -  & \textbf{\zz{2}} & \ding{56}& \textbf{\zz{0}} & -\\
ALU               & \textbf{\zz{0}} & - & \textbf{\zz{5}} & \ding{56}  & \textbf{\zz{0}}  & -  & \textbf{\zz{0}}  & -  & \textbf{\zz{0}} &-& \textbf{\zz{5}} & \ding{52}\\
PE                & \textbf{\zz{4}} & \ding{52}  & \textbf{\zz{5}} & \ding{52}  & \textbf{\zz{5}}  &  \ding{56} &  \textbf{\zz{5}} & \ding{52}  & \textbf{\zz{5}} & \ding{52}& \textbf{\zz{4}} & \ding{52}\\
risc\_cpu         & \textbf{\zz{0}} & -  &  \textbf{\zz{0}} &  - & \textbf{\zz{0}}  & -  &  \textbf{\zz{0}} & - & \textbf{\zz{0}} & -& \textbf{\zz{0}} & -\\

\midrule[0.6pt]
Success rate    & 55\%  & 10/30& 81\%  & 15/30  & 40\%  & 5/30  & 27\%  &   5/30  & 73\%  & 14/30& 90\% &19/30\\
\bottomrule[1pt]
\end{tabular}
}
\vspace{-.25in}
\end{center}
\end{table*}

Our proposed self-planning prompt engineering technique is actually similar to how we human beings solve a challenging task, like taking an exam or writing a complex algorithm. When we make a good plan ourselves before actually starting, we tend to perform better. Such similarity may imply certain humanlike ``intelligence'' of existing LLMs and can inspire further prompt engineering techniques in the future.



%% file: _txt/5_experiment.tex
\section{Experimental Results}


\subsection{Experiment Setup}
Given a design RTL, the design quality can be evaluated using synthesis tools. We perform logic synthesis with Synopsys Design Compiler\textsuperscript{\textregistered}~\cite{design-compilier}, using the advanced `compile\_ultra' synthesis option. We set the frequency to be extremely large to ensure a negative slack in all designs for an easier timing comparison. For functionality verification, the RTL simulation is performed with Synopsys VCS\textsuperscript{\textregistered}. 

In the experiment, we evaluated five LLMs with our proposed RTL generation benchmark: 
\begin{enumerate}
    \item GPT-3.5: the free commercial solution. 
    \item GPT-4.0: the state-of-the-art commercial solution. 
    \item Thakur et al.~\cite{thakur2023benchmarking}: an academic model with 16 billion parameters developed by fine-tuning the CodeGen model~\cite{nijkamp2022codegen} with Verilog data.
    \item StarCoder~\cite{li2023starcoder}: a recent general academic model with 15 billion parameters for code generation, without being fine-tuned for Verilog.
    \item GPT-3.5 + self-planning: adopting our proposed self-planning technique when using GPT-3.5.
    \item GPT-4.0 + self-planning: adopting our proposed self-planning technique when using GPT-4.0.
\end{enumerate}

Since there can be randomness in many LLM's outputs, for each test design in RTLLM, we query each LLM five times in five parallel sessions, with exactly the same description $\mathcal{L}$, then collect all five outputs $\mathcal{V}$, which may be different from each other. In our experiment results, we will evaluate the correctness of all five outputs for each test case. There is \emph{no} extra fixing of any incorrect output by human engineers or another round of query to LLMs.

\subsection{RTL Generation Correctness}

Table~\ref{syntax_function} summarizes the quantitative evaluation of both syntax and functionality correctness of all five evaluated LLMs using RTLLM. The syntax part counts the number of generated design RTLs $\mathcal{V}$ with correct syntax, out of the five trials. Then the functionality part (i.e., Func.) will count a success \ding{52} as long as there is one generated RTL successfully passing the testbench $\mathcal{T}$, out of the ones already with correct syntax. 

According to Table~\ref{syntax_function}, GPT-4.0, the state-of-the-art commercial LLM, achieves the highest performance with $81\%$ correct syntax and $15/30$ correct functionalities. In comparison, the GPT-3.5 alone degrades to $55\%$ correct syntax and $10/30$ correct functionalities. After using our self-planning together with GPT-3.5, the performance rise back to $73\%$ and $14/30$, which is close to the GPT-4's performance. It clearly validates the effectiveness of the self-planning technique. 

In comparison, the academic LLMs perform significantly worse, with $40\%$ syntax for Thakur et al.~\cite{thakur2023benchmarking} and $27\%$ for StarCoder~\cite{li2023starcoder}, both with $5/30$ functionality correctness. 

As demonstrated in this design correctness example, using our proposed RTLLM, we can automatically evaluate the performance of all LLMs in design RTL generation. In summary, the performance rank is GPT-4 + self-planning $>$ GPT-4 $>$ GPT-3.5 + self-planning $>$ GPT-3.5 $>$ Thakur et al.~\cite{thakur2023benchmarking} $>=$ StarCoder~\cite{li2023starcoder}.

\linespread{1.2}
\begin{table*}[htbp]
\vspace{-.05in}
\caption{The Design Qualities of Gate-Level Netlist, Synthesized with Design Compiler}
\label{tbl:PPA}
\resizebox{\textwidth}{!}{
\begin{tabular}{l|ccc|ccc|ccc|ccc|ccc}
\toprule[1pt]
\multirow{2}{*}{\textbf{Design}} &
  \multicolumn{3}{c|}{\textbf{Designer Reference ($\mathcal{V}_H$})} &
  \multicolumn{3}{c|}{\textbf{ChatGPT-3.5}} &
  \multicolumn{3}{c|}{\textbf{ChatGPT-4.0}} &
  \multicolumn{3}{c|}{\textbf{Thakur et al.~\cite{thakur2023benchmarking}}} &
  \multicolumn{3}{c}{\textbf{GPT-3.5 + Self-planning}} \\ \cline{2-16} 
 &
  \textbf{\begin{tabular}[c]{@{}c@{}}Area\\ ($\mu m^{2}$)\end{tabular}} &
  \textbf{\begin{tabular}[c]{@{}c@{}}Power\\ ($\mu W$)\end{tabular}} &
  \textbf{\begin{tabular}[c]{@{}c@{}}Timing\\ ($ns$)\end{tabular}} &
  \textbf{\begin{tabular}[c]{@{}c@{}}Area\\ ($\mu m^{2}$)\end{tabular}} &
  \textbf{\begin{tabular}[c]{@{}c@{}}Power\\ ($\mu W$)\end{tabular}} &
  \textbf{\begin{tabular}[c]{@{}c@{}}Timing\\ ($ns$)\end{tabular}} &
  \textbf{\begin{tabular}[c]{@{}c@{}}Area\\ ($\mu m^{2}$)\end{tabular}} &
  \textbf{\begin{tabular}[c]{@{}c@{}}Power\\ ($\mu W$)\end{tabular}} &
  \textbf{\begin{tabular}[c]{@{}c@{}}Timing\\ ($ns$)\end{tabular}} &
  \textbf{\begin{tabular}[c]{@{}c@{}}Area\\ ($\mu m^{2}$)\end{tabular}} &
  \textbf{\begin{tabular}[c]{@{}c@{}}Power\\ ($\mu W$)\end{tabular}} &
  \textbf{\begin{tabular}[c]{@{}c@{}}Timing\\ ($ns$)\end{tabular}} &
  \textbf{\begin{tabular}[c]{@{}c@{}}Area\\ ($\mu m^{2}$)\end{tabular}} &
  \textbf{\begin{tabular}[c]{@{}c@{}}Power\\ ($\mu W$)\end{tabular}} &
  \textbf{\begin{tabular}[c]{@{}c@{}}Timing\\ ($ns$)\end{tabular}} \\ \hline
accu & 239 & 19K & -0.42 & 298 & 24K & -0.43 & 304 & 21K & -0.39 & - & - & - &\correct{231}  & \correct{18K} & \correct{-0.37}  \\
adder\_8bit & 65 & 34 & -0.62 & 38 & 14 &-0.14& \correct{15} & \correct{5.8} & \correct{-0.12} & - & - & - & 74 &42  & -0.63 \\
adder\_16bit & 128 & \correct{68} & -1.21 & \wrong{157} & \wrong{91.0} & \wrong{-0.33} & \correct{126} & \correct{68} & -1.19 & 189 & 106 &-0.31  & 163 &94  & \correct{-0.33} \\
adder\_32bit & 571 & 298 & -0.72 & \wrong{58} & \wrong{17} & \wrong{-0.04} & \wrong{65} & \wrong{26} & \wrong{-0.13} & - & - &-  & \correct{337} & \correct{199} &\correct{-0.43}  \\
adder\_64bit & 2.9K & 296K & -0.48 & \wrong{2.5K} & \wrong{242K} & \wrong{-0.60} & \wrong{2.4K} & \wrong{187K} & \wrong{-0.48} &-  & - & - & \wrong{2.3K} & \wrong{220K}  & \wrong{-0.32}  \\
multi\_8bit & 52 & 6.1K & -0.08 & \wrong{640} & \wrong{45K} & \wrong{-0.43} & \wrong{494} & \wrong{33K} & \wrong{-0.49} &-  & - &-  & \wrong{259} & \wrong{23K} & \wrong{-0.27} \\
multi\_16bit & 749 & \correct{75K} & -0.91 & - & - & - & \correct{531} & 79K & \correct{-0.50} & 7.5K & 384K &-1.76  & -  & -  & - \\
multi\_pipe\_4bit & 198 & \correct{19K} & -0.34 & - & - & - & \correct{193} & 22K & \correct{-0.33} & - & - & - & \wrong{146} & \wrong{17K} & \wrong{-0.30} \\
multi\_pipe\_8bit & 961 & 78K & -0.65 & - & - & - & \wrong{1.1K} & \wrong{80K} & \wrong{-0.99} &-  & - & -& \wrong{443} & \wrong{42K} &\wrong{-0.14} \\
div\_8bit & 158 & 8.4K & -0.38 & - & - & - & - & - & - & - & - &-  & - & - &-  \\
div\_16bit & 1.8K & 2.4K & -4.20 & - & - & - & \wrong{1.5K} &\wrong{ 1.8K} & \wrong{-4.84} & - & - &-  &-  & - &-  \\
JC\_counter & 380 & 45K & \correct{-0.13} & 380 & 45K & \correct{-0.13} & \correct{42} & \correct{4.7K} & -0.26 &\wrong{29}  &\wrong{4.6K}  & \wrong{-0.23} & 195 & 21K & -0.22 \\
right\_shifter & 42 & 4.2 & -0.14 & \correct{40} & \correct{3.8K} & \correct{-0.12} & 46 & 5.7K & -0.13 & \correct{40}  &\correct{3.8K}  & \correct{-0.12}  & \correct{40}  &\correct{3.8K}  &  \correct{-0.12} \\
mux & 68 & \correct{6.5} & \correct{-0.08} & - & - & - & 90 & 9.5 & \correct{-0.08} &\correct{64}  & 13 &  \correct{-0.08} & \wrong{144} & \wrong{14}  &\wrong{-0.08} \\
counter\_12 & 49 & \correct{4.3K} & -0.31 & 79.0 & 8.0K & \correct{-0.25} & \correct{46} & 4.4K & -0.26 & \wrong{35}&\wrong{4.0K}  & \wrong{-0.24}  &76  & 8.4K & -0.26 \\
freq\_div & 124 & 16K & -0.29 & \wrong{911} & \wrong{66K} & \wrong{-0.45} & \wrong{118} & \wrong{16K} & \wrong{-0.32} &\wrong{226}  &\wrong{16K}  &  \wrong{-0.4} & \wrong{667} & \wrong{53K} & \wrong{-0.41} \\
signal\_generator & 178 & 14K & -0.36 & \correct{72} & \correct{9.2K} & \correct{-0.23} & 98 & 11K & -0.26 &-  &  -& - & 101 & 11K & -0.27 \\
serial2parallel & 135 & 13K & -0.29 & 168 & 16K & -0.30 & \correct{100} & \correct{9.8K} & \correct{-0.28} & - & - & - &155  & 14K &-0.33  \\
parallel2serial & 55 & 8.6K & -0.23 & \wrong{35} &\wrong{ 6.2K} & \wrong{-0.21} & \wrong{20} & \wrong{3.8K} & \wrong{-0.19} & - & - &-  & \wrong{1.06} & \wrong{0} & \wrong{0} \\
pulse\_detect & 25 & 2.8 & -0.13 & \wrong{42} & \wrong{2.8} & \wrong{-0.12} & \wrong{40} & \wrong{4.3} & \wrong{-0.08} & \wrong{25} & \wrong{2.8} &\wrong{-0.12}  &\wrong{28}  & \wrong{3.4} & \wrong{-0.08} \\
edge\_detect & \correct{19} & \correct{2.6K} & \correct{-0.14} & 24 & 3.3K & -0.16 & \correct{19} & \correct{2.6K} & \correct{-0.14} & \wrong{1.06} & \wrong{0} & \wrong{0} & \correct{19} & \correct{2.6K} & \correct{-0.14}  \\
FSM & 44 & 3.5K & -0.18 & \wrong{26} &\wrong{ 2.7K} & \wrong{-0.21} & \wrong{34} & \wrong{2.7K }& \wrong{-0.25} &\wrong{27}  & \wrong{2.7K} &\wrong{-0.24 } &  \wrong{45}& \wrong{4.1K} & \wrong{-0.2} \\
width\_8to16 & 219 & 23K & -0.26 & 214 & 21K & \correct{-0.20} & 219 & 23K & -0.26 & - & - & - & \correct{144} &\correct{14K}  &0.24  \\
traffic\_light & 178 & 18K & -0.35 & \wrong{147} & \wrong{14K} & \wrong{-0.34} & 138 & 11K & -0.38 & - & - & - & - & - & - \\
calendar & \correct{199} & \correct{16K} & \correct{-0.36} & - & - & - & \wrong{460} &\wrong{31K} & \wrong{-0.51} & - & - & - &227  & \correct{16K} &-0.37  \\
RAM & 3.5K & 248K & -0.35 & - & - & - & - & - & - & 353 & 27K & -0.26 & - & - & - \\
asyn\_fifo & 1.3K & 107 & -0.23 & - & - & - & - & - & - & - & - & - & \wrong{0} & \wrong{0} & \wrong{0} \\
ALU & 2.4K & 1.0K & -0.76 & - & - & - & \wrong{3.3K} &\wrong{ 1.4K} & \wrong{-0.71} & - & - & - & - &-  &-  \\
PE & \correct{2.4K} & 363K & \correct{-1.03} & 2.5K & 359K & -1.08 & 2.6K & 366K & -1.06 &\wrong{ 2.2K }& \wrong{275K} & \wrong{-0.07} &2.5K  & \correct{358K} &-1.08  \\
risc\_cpu & 634 & 6.2K & -0.30 & - & - & - & - & - & - & - & - & - &-  &-  & - \\  
\hline
Best Quality Num &  3 & 7 & 5  & 2 & 2 & 5 &  8 & 5 & 6 &  2 & 1 & 2 & 5 &  7 & 5 \\
\bottomrule[1pt]
\end{tabular}
}
\vspace{-.15in}
\end{table*}

\subsection{RTL Generation Quality}

After evaluating design correctness, our RTLLM further supports evaluating the design qualities in power, timing, and area. Table~\ref{tbl:PPA} summarizes the design qualities of generated design RTL from different LLMs\footnote{The worst LLM StarCoder is not presented due to space limitation.}. These quality values are measured on each post-synthesis netlist. We report the worst negative slack (WNS) as the timing metric. It also presents the qualities of our designer-generated reference design $\mathcal{V}_H$ in RTLLM. All these reference designs are functionally correct. 

For each generated design RTL $\mathcal{V}$, as long as it can be correct in syntax, we can perform the logic synthesis and report its design qualities in Table~\ref{tbl:PPA}. We then mark the design RTLs with correct syntax but wrong functionality (i.e., fail to pass testbench) in Table~\ref{tbl:PPA} as red color. Those un-synthesizable designs with the wrong syntax are left blank in Table~\ref{tbl:PPA}.  

For each design from RTLLM, we mark the generated design with the best power, performance, and area among all candidates in green color. Then we count the number of best qualities achieved by each LLM method. Of course, only designs that are both syntax and functionality correct are eligible for this comparison and can be colored green. 

According to the last row of Table~\ref{tbl:PPA}, the GPT-4 achieves the highest number of best qualities. GPT-3.5 + self-planning ranks second, with 5, 7, 5 designs achieving the best area, power, and timing, respectively. Both of them perform better than the designer-crafted reference designs $\mathcal{V}_H$. This trend of design quality is similar to the trend of design correctness, indicating GPT-4.0 $>$ GPT-3.5 + self-planning $>$ GPT-3.5 $>$ Thakur et al.~\cite{thakur2023benchmarking}. Please notice that, since there is a strong trade-off between different design objectives, this summation of individual best design quality leads to a straightforward but less rigorous comparison.


%% file: _txt/6_conclusion.tex
\section{Conclusion}

In this work, we propose a comprehensive open-source benchmark for design RTL generation with natural language instructions. Compared with the datasets released in recent works, our benchmark includes more designs, also with higher design scale and complexity. We also propose an effective prompt engineering technique named self-planning. In our future work, we will first keep extending and maintaining this benchmark. We will also keep validating the self-planning technique. In addition, we will fine-tune our own open-source models to achieve better performance in our RTLLM benchmark.

\section*{Acknowledgement}

This work is partially supported by the National Natural Science Foundation of China (92364102, 62304192), Hong Kong Research Grants Council (RGC) ECS Grant 26208723, and ACCESS – AI Chip Center for Emerging Smart Systems, sponsored by InnoHK funding, Hong Kong SAR.

